\documentclass[conference]{IEEEtran}
\makeatletter
\def\ps@headings{%
\def\@oddhead{\mbox{}\scriptsize\rightmark \hfil \thepage}%
\def\@evenhead{\scriptsize\thepage \hfil \leftmark\mbox{}}%
\def\@oddfoot{}%
\def\@evenfoot{}}
\makeatother
\usepackage{amsmath,amssymb,amsfonts}
\DeclareMathOperator*{\argmax}{arg\,max}

\usepackage{algorithm}
\usepackage{algpseudocode}
\usepackage{textcomp}
\usepackage{xcolor}

\usepackage{array}
\usepackage{multirow}
\usepackage{booktabs}
\usepackage{enumitem}
\usepackage[flushleft]{threeparttable}

\usepackage{graphicx}
\usepackage{xcolor}

\usepackage{cite}
\usepackage{hyperref}
\hypersetup{
    colorlinks=true,
    linkcolor=blue,
    filecolor=magenta,      
    urlcolor=cyan,
}

\definecolor{darkred}{rgb}{0.8,0,0}
\definecolor{darkgreen}{rgb}{0,0.5,0}
\newcommand{\tit}[1]{\smallbreak\noindent\textbf{#1 }}
\newcommand{\low}{{\color{darkred} $\downarrow$}}
\newcommand{\up}{{\color{darkgreen} $\uparrow$}}

\def\BibTeX{{\rm B\kern-.05em{\sc i\kern-.025em b}\kern-.08em
    T\kern-.1667em\lower.7ex\hbox{E}\kern-.125emX}}
\begin{document}

\title{Are Existing Out-Of-Distribution Techniques Suitable for Network Intrusion Detection?}

\author{
\IEEEauthorblockN{Andrea Corsini}
\IEEEauthorblockA{\textit{Department of Science and Methods for Engineering} \\
\textit{University of Modena and Reggio Emilia}, Modena, Italy \\
andrea.corsini@unimore.it}
\and
\IEEEauthorblockN{Shanchieh Jay Yang}
\IEEEauthorblockA{\textit{Department of Computer Engineering} \\
\textit{Rochester Institute of Technology}, Rochester, USA \\
jay.yang@rit.edu}
}
\maketitle

\begin{abstract}
Machine learning (ML) has become increasingly popular in network intrusion detection. However, ML-based solutions always respond regardless of whether the input data reflects known patterns, a common issue across safety-critical applications.
While several proposals exist for detecting Out-Of-Distribution (OOD) in other fields, it remains unclear whether these approaches can effectively identify new forms of intrusions for network security.
New attacks, not necessarily affecting overall distributions, are not guaranteed to be clearly OOD as instead, images depicting new classes are in computer vision.
In this work, we investigate whether existing OOD detectors from other fields allow the identification of unknown malicious traffic.
We also explore whether more discriminative and semantically richer embedding spaces within models, such as those created with contrastive learning and multi-class tasks, benefit detection.
Our investigation covers a set of six OOD techniques that employ different detection strategies. These techniques are applied to models trained in various ways and subsequently exposed to unknown malicious traffic from the same and different datasets (network environments).
Our findings suggest that existing detectors can identify a consistent portion of new malicious traffic, and that improved embedding spaces enhance detection.
We also demonstrate that simple combinations of certain detectors can identify almost 100\% of malicious traffic in our tested scenarios.
\end{abstract}
\section{Introduction}\label{sec:intro}

Network Intrusion Detection Systems (NIDS) monitor the network traffic for signs of potential threats with various techniques, including signature detection, anomaly detection, and behavioral analysis~\cite{DataNIDS, survey2019}.
Network traffic can be analyzed either at a Packet Capture or Network Flow~\cite{flowIDS} (NetFlow) level, though packet inspection has become less common due to encryption and the massive size of modern traffic. 
We focus on NetFlow inspection, where packets relating to a single communication~\cite{flowIDS} are analyzed by measuring aggregated features such as idle times and the amount of exchanged data.

Recently, Machine Learning has gained popularity in NIDS~\cite{NIDSurvey} as it enables automatic extraction of complex detection patterns, quick adaptation to changing environments~\cite{L2Adapt}, and easy personalization without expensive human expertise.
However, ML-based solutions have limitations such as lacking interpretability and requiring well-crafted training data.
Although a large body of research is addressing these issues~\cite{NIDSurvey}, we focus on another drawback: \emph{ML-based NIDSs always provide a response regardless of whether they recognize (are trained with) the input data pattern}.

This issue is particularly relevant since network traffic tends to have \emph{dynamic and non-stationary distributions}, either caused by normal behavioral shifts or adversaries, which can cause degradation in NIDS performance: a problem known as \emph{concept drift}~\cite{ConceptDrift}.
Another inherent problem of dynamic distributions is Out-Of-Distribution data, which is \emph{unusual traffic markedly different from a reference distribution} not necessarily affecting the overall data distribution.
In general, concept drift and OOD data are both caused by shifts in feature distributions, label distributions, or both~\cite{ConceptDrift,generalizedOOD}.

\begin{figure}[t]
    \centering
    \includegraphics[width=.9\columnwidth]{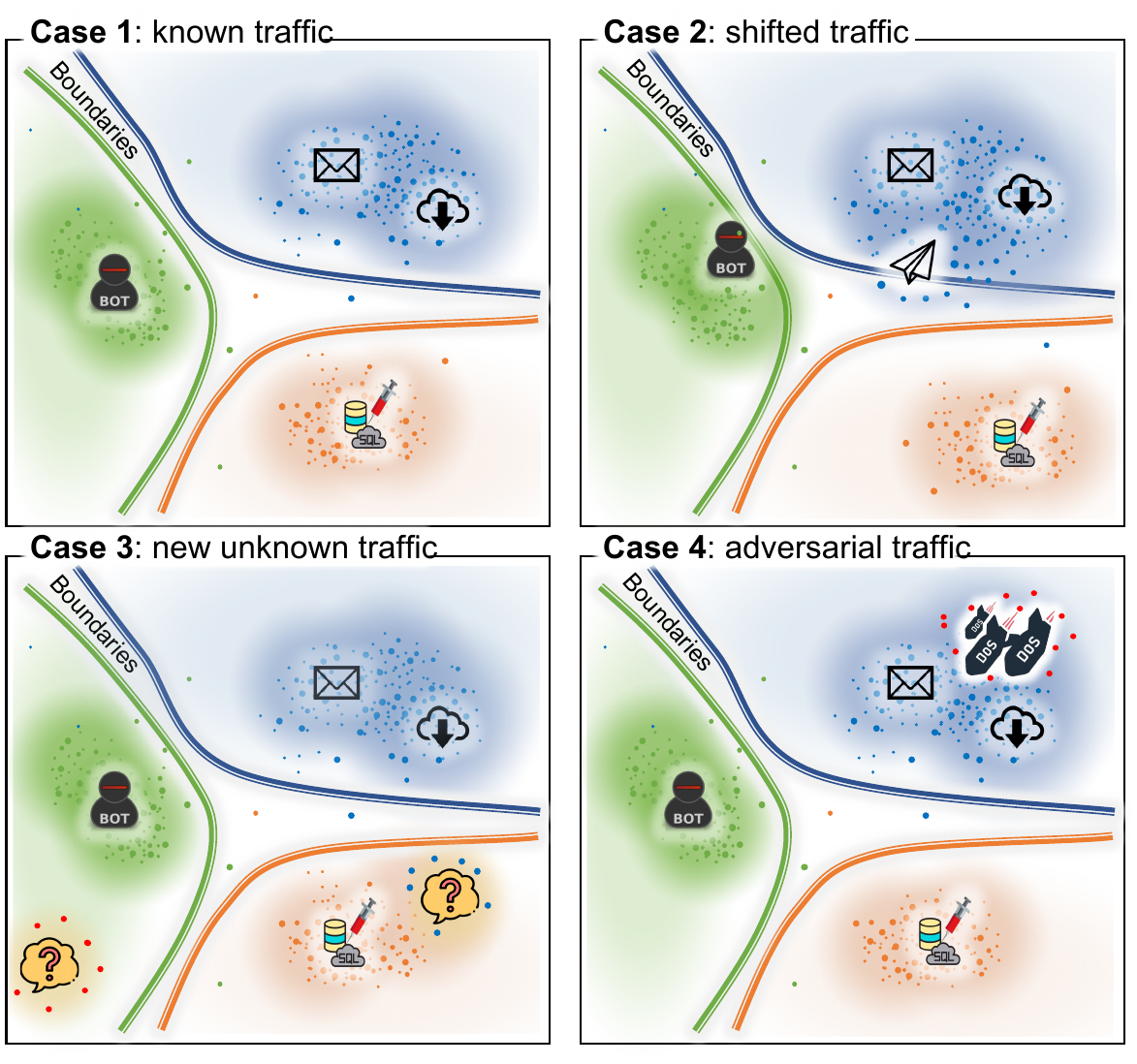}
    \caption{Different situations in the decision space of a deployed ML-based NIDS. Case 1 is the expected situation where the new traffic respects the i.i.d. assumption. Case 2 depicts traffic that is (gradually) shifting due to changes in malicious and benign behaviors towards relatively known (usual) regions. Case 3 shows new unknown traffic (of potentially different classes) falling into unusual regions of the decision space. Case 4 describes a challenging situation where a new attack is crafted so that it is misclassified by the NIDS.}
    \label{fig:spots}
\end{figure}
Based on our analysis, an ML-based NIDS may be affected in various ways after deployment.
In Fig.~\ref{fig:spots}, we present 4 exemplar cases in a NIDS trained to detect Botnet and SQL injections.
Normally, the NIDS is expected to work as in Case 1, where new traffic is well represented by training data (i.i.d. assumption). 
However, it is normal for traffic to shift over time and this may affect NIDSs depending on the shift extent and direction.
For instance, the shift to SQL traffic in Case 2 does not compromise the NIDS, but the same is not true for benign traffic.
Moreover, new traffic may also be adversarially crafted (Case 3) and be potentially mistaken as in Case 4.

We argue that shifts as those in Fig.~\ref{fig:spots} happen more in NetFlow features (\emph{covariate shift}~\cite{CovariateShift,generalizedOOD}) rather than in labels (\emph{actual or semantic shift}~\cite{ConceptDrift,generalizedOOD}).
As in Case 2, malicious traffic remains malicious if its features are adversarially crafted to evade detection, while shifts in normal user behaviors should not transform benign traffic into malicious.
Even in cases of new attacks (Case 3), it might be possible to experience shifts in feature distributions~\cite{generalizedOOD}.
Additionally, we argue that OOD techniques sensitive to small feature perturbations can also serve as drift detectors by monitoring the volume of alerts over time either with standard statistics or existing methods~\cite{adwin,ConceptDrift,CovariateShift}.
Therefore, in this work, we adapt and evaluate techniques to detect shifts primarily affecting features.

A perfect OOD detector should trigger an alert and ask the expert knowledge for further investigation in every situation but Case 1.
However, Case 4 is extremely difficult to detect without any additional information besides the ML-based NIDS and training data.
Whereas well-designed OOD detectors should identify cases like 2 and 3.
We thus investigate whether traffic generated by new attacks, either similar to those in training or completely different, can be detected as OOD by existing techniques from other ML fields. 
Note that it is not guaranteed that effective techniques in other fields are suitable for network intrusion.
As an example, well-working detectors on data like images with bounded and discrete domains might prove ineffective on NetFlows, where features are generally a mix of continuous unbounded and discrete.

Therefore, we select detection techniques of different natures from other ML fields and evaluate whether such techniques can identify NetFlows of unknown attacks.
As baseline model, we consider a standard FeedForward Neural Network, and we also assess the effect of different training regimes on the quality of detection techniques.
Specifically, we train models in \emph{binary} (different attacks in the same class) and \emph{multi-class} (each attack makes a class) settings, with and without the aid of a simple Contrastive Learning approach: \emph{Center-Loss}~\cite{centerloss}.
We expose various combinations of models and detection techniques to malicious traffic generated from attack types not seen in training, where such traffic may come from the same dataset (same network environment) and a different dataset (different network environment).
Finally, we evaluate two ensembles of OOD techniques to enhance detection and further explore the complementarity of these techniques, providing guidelines for practical applications.
All our code and the numerical results are freely available at \url{https://github.com/AndreaCorsini1/CyberOOD}

The contributions of this paper include:
\begin{itemize}
    \item We investigate the effectiveness of treating the identification of unknown intrusions as an OOD detection problem and explore the applicability of existing OOD techniques.
    \item We identify the most effective techniques for detecting new intrusions and explore their potential for combination to enhance detection. We also discuss limitations of some techniques, providing insights for further development.
    \item We emphasize the significance of improving the model embeddings to achieve better detection, highlighting that:
    \begin{itemize}
        \item \emph{Contrastive Learning}, specifically the use of \emph{Center-Loss}, enables the creation of embeddings that improve OOD techniques and their ensemble.
        \item \emph{Multi-class} training allows making semantically richer embeddings, which offer advantages over binary ones for OOD techniques and their ensembles.
    \end{itemize}
\end{itemize}

The remainder is organized as follows: Sec.~\ref{sec:related} presents existing OOD literature; Sec.~\ref{sec:back} describes key concepts for our work; Sec.~\ref{sec:proposal} outlines our methodology; Sec.~\ref{sec:setup} describes the experimental setup; Sec.~\ref{sec:results} presents results; and Sec.~\ref{sec:conclusion} closes with limitations and potential future directions.
\section{Related Works}\label{sec:related}

In this section, we present various Out-Of-Distribution detection techniques and we review recent proposals to identify and react to shifts in the NIDS literature.

\subsection{Out-Of-Distribution in Machine Learning}\label{ssec:ood_ml}

Machine learning models are trained under the closed-world assumption, where test data is drawn i.i.d. from the same distribution as the training data.
However, this assumption is often violated and several ML fields try to address the issue of identifying unknown/anomalous/out-of-distribution data:
\begin{itemize}[leftmargin=3.5mm]
    \item \emph{Anomaly detection}~\cite{SurveyAnomaly} aims to detect anomalous inputs that deviate from normality, whether in features or labels. Anomaly detection assumes there might be abnormal data in the training set~\cite{unifiedOOD} and treat data as a whole, thus it does not strictly require the correct classification of inputs.
    \item \emph{Novelty detection} is similar to anomaly detection, but assumes the presence of only normal data in the training set and focuses on inputs affected by semantic shift~\cite{generalizedOOD}, hence not falling into any of the training classes. In addition, novel inputs are not treated as erroneous and are typically prepared for retraining and future constructive procedures.
    \item \emph{Open Set Recognition}~\cite{OSR} goes beyond novelty detection and also requires the correct classification of in-distribution (ID) data. The goal is to detect inputs belonging to new classes and correctly classify those from known classes. Open Set Recognition is usually focused on semantic shifts. 
    \item \emph{Outlier detection} identifies inputs in a dataset that markedly differ from others. Outlier detection is a pre-processing step and is not applied during inference or training.
\end{itemize}

As introduced in Sec.~\ref{sec:intro}, the NIDS setting requires the classification of known traffic and the detection of shifts caused either by modifications in known traffic or the appearance of unknown traffic.
This setting resembles the Open Set Recognition one, but it additionally comprises shifts not implying the appearance of new classes.
Therefore, we speak of Out-Of-Distribution detection in general terms.

\tit{Confidence-based}detectors use estimates derived from a model to quantify the level of certainty or trust in its predictions as an indicator of ID-ness.
In~\cite{confOOD}, the authors observed that well-trained models assign lower confidence scores to OOD data.
Subsequent studies~\cite{ODIN,React,MCDropout} have proposed techniques to enhance confidence estimation, while others have introduced modifications to the model architecture and training objectives~\cite{LearningConfidence,GenODIN}.
Although confidence is not always a reliable OOD indicator~\cite{notConf,EnoughUQ}, due to their simplicity and clarity, confidence-based detectors are commonly used in practice and serve as a baseline for OOD detection.

\tit{Density-based}detectors explicitly model the distribution of ID data, either raw or latent features, and flag samples falling into low-density regions as OOD. 
In multi-class tasks, class-conditional distribution estimators are often employed so that the OOD samples can be identified based on their likelihood~\cite{DensityOOD,DensityRegret}.
To model the class-conditional distribution of ID data, parametric and non-parametric models such as a simple Mixture of Gaussian, Kernel Density Estimation, and deep generative models~\cite{generalizedOOD} are frequently used.
However, modeling the distribution of complex data and estimating the likelihood may be challenging~\cite{DensityRegret}, imply a-priori assumptions that need validation, and do not always scale well like in kernel estimators. 
Therefore, we prefer to avoid these detectors and leave their evaluation to future work.

\tit{Distance-based}detectors are based on the idea that the OOD samples should be relatively far away from centroids or prototypes of ID classes. 
Once a prototype is extracted for each training class, a distance metric like Mahalanobis, Euclidean, or Cosine can be used to estimate the class similarity and flag samples that are not similar enough to any of the prototypes~\cite{DensityOOD,generalizedOOD}.
Recently, even a class-conditioned K-Nearest Neighbor approach~\cite{knnOOD} has been adopted to detect OOD samples based on the distance from the k-nearest neighbor.

\subsection{Out-Of-Distribution in Network Intrusion}\label{ssec:ood_nids}

A significant portion of the network intrusion literature on ML applications focuses on anomaly detection~\cite{SurveyNIDSAnomaly,bat} and concept drift~\cite{ConceptDrift,cade,ENIDrift,INSOMNIA}.
Anomaly techniques, such as autoencoders~\cite{deeplearning}, have gained interest due to their ability to detect unknown attacks using only normal traffic and without requiring labels. 
However, these methods often suffer from a high number of false alarms as they flag any anomalous sample as an attack~\cite{NIDSurvey}.
In contrast, concept drift and OOD detectors are generally more effective but typically require labeled data~\cite{ConceptDrift,generalizedOOD}.
Therefore, recent works proposed ML-based solutions that ease the need for labels without increasing false alarms by leveraging anomaly detection techniques.
For instance, \cite{kitsune} proposed an efficient and online ensemble of autoencoders that utilizes an ad-hoc feature extraction module to differentiate normal and abnormal patterns in packets. 
Similarly, \cite{kitsune} introduced an adaptive ensemble system that incorporates a packet-based feature extraction method and a sub-classifier generation module to create ensemble models from drifted data chunks and ground truth labels. 
\cite{ConceptEnsemble} modified the extreme gradient boosting algorithm to detect and adapt to drifts in the presence of a large number of features. 
\cite{INSOMNIA} employed active learning, label estimation, and an explainable ML framework to respectively update the model, reduce labeling overhead, and interpret model reactions to shifts.

In a context akin to ours, \cite{L2Adapt} utilized a contrastive loss signal alongside a distance function capturing instance and class-level fidelity to recursively update the encoder network.
Similarly, \cite{cade} employed Contrastive Learning to create a compressed representation of training data which is used to detect drifting samples with class centroids.
Both these works use a contrastive signal that pulls embeddings of the same class together and pushes those of different classes apart, we instead rely on Center-Loss~\cite{centerloss}.
Moreover, these works use autoencoders while we adopt a FeedForward Network.
\section{Gradient Detection \& Contrastive Learning}\label{sec:back}

\begin{figure*}[t]
    \centering
    \includegraphics[width=0.8\textwidth]{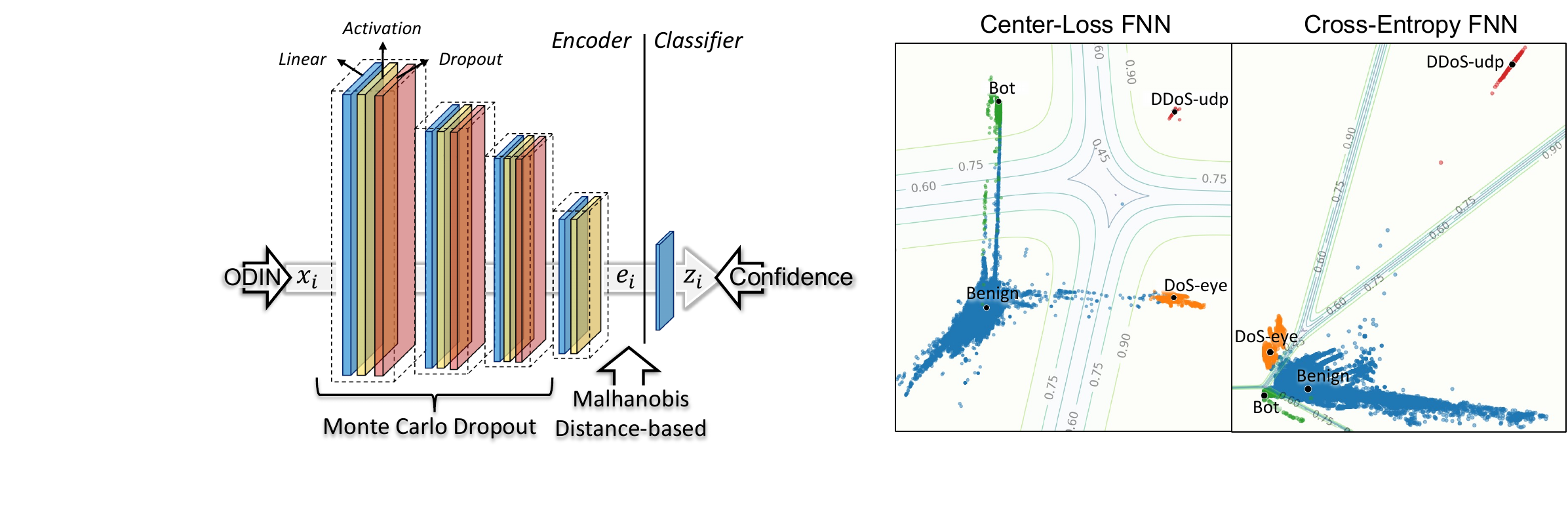}
    \caption{On the left, the architecture of the considered FNN and the holistic view of where the different OOD detectors act. On the right, the 2D embeddings ($e_i$) created by the encoder inside the decision space of the FNN trained with and without Center-Loss on four traffic types. The lines highlight points of the decision space where the softmax scores produced by the classifier (i.e., the FNN confidence) change.}
    \label{fig:tech}
\end{figure*}
This section presents key components of our study, specifically gradient-based detectors and Contrastive Learning.
We represent an ML-based NIDS with a parameterized model $f$ that maps NetFlows $x_i \in \mathbb{R}^d$ into a class $\overline{y} = \argmax_{j \in C} z_j$, where $C$ is the set of training classes and $z_j = f_j(x_i)$ is the logit (pre-softmax) score produced by $f$ for class $j \in C$.
Additionally, we suppose the model gives in output the embedded representation $e_i \in \mathbb{R}^w$ of $x_i$ constructed after the last embedding layer, i.e., the one before the classifier layer.
Refer to the left part of Fig.~\ref{fig:tech} for a graphical representation.

\subsection{Gradient-based Detection: ODIN and Mahalanobis}\label{ssec:gradient}

Most OOD detectors rely on information extracted from models to derive OOD scores, disregarding information on the gradient.
In \cite{ODIN}, the authors observed that adding a fixed perturbation to samples in the direction of the gradient amplifies the gap between ID and OOD softmax scores.
Thus, the idea behind Out-of-DIstribution detectioN (ODIN)~\cite{ODIN} is to jointly apply \emph{temperature scaling}~\cite{Calibration} and a \emph{controlled perturbation} to detect OOD data. 
ODIN consists of the following steps:
\begin{enumerate}
    \item \textbf{Temperature Scaling}: divides the logits $z_j$ by a temperature $T$ that reduces the sharpness of the softmax distribution and makes the model less confident.
    \item \textbf{Perturbation}: involves adding a perturbation $\epsilon$ to $x_i$ in the direction given by the sign of the gradient:
    \begin{equation}
        \widehat{x_i} = x_i - \epsilon \text{sign}(\nabla p_j^*)
    \end{equation}
    where $p_j^* = \max_{j \in C} p_j$ is the maximum softmax score for $x_i$ after temperature scaling. This perturbation pushes samples toward their nearest class.
    \item \textbf{Detection}: computes an ID score by feeding $\widehat{x_i}$ into the model again; if this score is above a threshold, $x_i$ is ID.
\end{enumerate}

Another similar gradient-based method is the Mahalanobis Detector (MD)~\cite{unifiedOOD}, where the Mahalanobis distance is used to measure how ``typical'' a point is with respect to a learned latent distribution. 
The Mahalanobis distance requires an estimate of the mean $\mu$ and covariance matrix $\Sigma$ for each ID class, which are normally extracted from the training set. 
After having these parameters, MD applies a controlled perturbation as in ODIN, but without temperature scaling and where the gradient is computed with respect to the distance between $e_i$ and the nearest class distribution ($dist_{MD}(\cdot)$):
\begin{equation}
    \widehat{x_i} = x_i - \epsilon \text{sign}(\nabla dist_{MD}(e_i))
\end{equation}

The final OOD detection is similar to ODIN: a threshold is first extracted from the validation, and every perturbed $\widehat{x_i}$ with a distance higher than this threshold is labeled as OOD.

\subsection{Contrastive Learning and Center Loss}\label{ssec:contrastive}

Contrastive Learning is a self-supervised technique designed to learn meaningful embedding representations. 
It achieves this by bringing similar input samples closer together in the learned embedding space while pushing dissimilar apart~\cite{constrastive}. 
By doing so, Contrastive Learning encourages the model to capture discriminative features that can be useful for various downstream tasks.
In a typical contrastive framework, each sample in a batch is augmented through ad-hoc transformations (such as random cropping and flipping for images) into new samples called the positives, while the original sample is referred to as the anchor.
The objective is to maximize the similarity between the anchor and the positives while minimizing the similarity between the anchor and other batch samples.

One of the precursor techniques to Contrastive Learning is Center-Loss~\cite{centerloss} (CL).
CL encourages a model to learn discriminative embeddings $e_i$ that cluster around their class centers.
It accomplishes this by defining a center $c_j \in \mathbb{R}^w$ for each class $j \in C$ and introducing an additional term to the standard cross-entropy loss. 
This additional term minimizes the distance between the embeddings and their corresponding class centers, which are determined by the ground-truth labels. 
The class centers are learned alongside the model's parameters by minimizing the additional Center-Loss term:
\begin{equation}\label{eq:cl}
    L_{CL} = \frac{1}{2} \sum_{i \in B} ||e_i - c_j^*||^2
\end{equation}
where the sum is over the batch samples $B$ and $c_j^*$ is the ground-truth center of each input.
The overall loss is thus a linear combination of Cross-Entropy ($L_{CE}$) and Center-Loss: $L = L_{CE} + \lambda L_{CL}$, where $\lambda$ is a hyperparameter that controls the weight of $L_{CL}$.
\section{Methodology \& Design Choices}\label{sec:proposal}

Herein, we present our model architecture, the adapted Center-Loss for our settings, and the selected OOD detectors.

\subsection{The Model Architecture}\label{ssec:model}

Although it might be possible to design ad-hoc architectures for OOD detection tasks~\cite{LearningRejection,GenODIN}, we prefer to avoid them and make no particular assumption about the model.
We only require for a NetFlow $x_i \in \mathbb{R}^d$ to have access to its pre-softmax score $z_i$ and to an embedded representation $e_i \in \mathbb{R}^w$ produced within the model, like the one generated before the classification layer.
Therefore, the architecture can comprise any layer like convolutional, linear, and recurrent ones~\cite{deeplearning}.

We logically divide our model into two parts: (i) an encoder that transforms NetFlows $x_i$ into embeddings $e_i$, and (ii) a classifier that uses $e_i$ to produce a softmax score for each class.
The proposed encoder is composed of four linear layers of decreasing size, each activated through a LeakyReLU non-linearity with a slope of 0.15.
We also apply dropout after the first three layers.
The classifier is a single linear layer that has as many neurons as the number of output classes.
We will refer to such a model as Feedforward Neural Network (FNN) and provide in the left part of Fig.~\ref{fig:tech} a visual representation.

\subsection{Improving the Model Embedding}\label{ssec:train}

Recently, Contrastive Learning has been widely adopted to improve the performance of ML in different tasks~\cite{constrastive}.
In our work, we propose to use Contrastive Learning to make embeddings learned by our FNN more discriminative, improving classification tasks~\cite{constrastive,centerloss,supcon} and potentially enhancing the effectiveness of OOD detectors.
As an example, refer to the two plots on the right of Fig.~\ref{fig:tech}, which represent the embedding spaces produced by our FNN encoder when trained with and without a contrastive learning signal, respectively.
It is immediate to see that the projected NetFlows of individual attacks are less scattered and more separated in the Center-Loss plot.
These discriminative embeddings may benefit detectors like distance-based ones that assume normality or well-representative prototypes to detect OOD.

Although many contrastive methods exist~\cite{supcon,centerloss,constrastive}, most of them are primarily designed for other ML fields and rely on positive samples and ad-hoc augmentations~\cite{constrastive}, vague concepts in the NIDS literature.
Therefore, we prefer to employ a simpler and more straightforward method: \emph{Center-Loss} (described in Section~\ref{ssec:contrastive}).
The application of Center-Loss to our setting does not require any particular modifications; however, we need to account for the unique aspects of NIDSs such as heavily imbalanced training sets and noisy data.

To mitigate the effect of imbalanced sets, we adopt a combination of over- and under-sampling as further described in Sec.~\ref{ssec:details}.
This is particularly important because CL works locally on batches, and with heavy unbalancing it is likely to have batches with only NetFlows of the majority benign class, thus focusing too much on improving their embeddings and its center.
In addition, we propose to apply the CL term of Eq.~\ref{eq:cl} only on samples correctly classified by the model.
This helps in mitigating the effect of noisy labels and data during training, which are common issues in NIDSs~\cite{issues2017}.

\subsection{Adopted OOD Detectors}\label{ssec:detectors}

In this work, we consider OOD detectors of different natures that work beside classification models (pre-trained and not) and can be applied to any architecture. 
Our rationale for selecting detectors reviewed in Sec.~\ref{sec:related} is to choose popular ones in related ML fields whose complexity (theoretical and implementation) is as low as possible.
Wherever possible and not penalizing in terms of performance, we prefer to evaluate detectors as originally proposed.

Regarding \underline{confidence-based} detectors, we adopt the baseline approach proposed in~\cite{BaselineOOD} (\texttt{CONF}).
This straightforward solution involves applying a threshold to the softmax scores and labeling as OOD all the NetFlows with a score below this threshold. 
We also adopt Monte Carlo Dropout~\cite{MCDropout} (\texttt{MCD}) in a similar manner. 
Instead of relying on a single confidence estimate for a NetFlow $x_i$, we leverage MCD with a switch-off probability of 0.4 to obtain multiple softmax scores.
Then, all those $x_i$ for which the standard deviation of their softmax scores exceeds a predefined threshold are flagged as OOD.
This allows a less biased estimate about $x_i$. 

In addition, we adopt two cutting-edge \underline{gradient-based} detectors in computer vision which are \texttt{ODIN}~\cite{ODIN} and Malahnobis~\cite{unifiedOOD} (\texttt{MD}), introduced in Sec.~\ref{ssec:gradient}.
Although there exist improvements over these proposals (see e.g.~\cite{GenODIN}), we prefer to keep them as originally proposed to avoid potential biases introduced by the assumptions of such improvements.
As we demonstrate later, gradient-based detection seems less effective on NetFlows compared to images.

Lastly, we include two \underline{distance-based} detectors.
The first one (\texttt{SIM}) uses the simplified Silhouette~\cite{silhouette} to measure the distance between test and training data.
For each class $j \in C$, \texttt{SIM} first extracts a center by averaging the embeddings $e_i$ of training data labeled with $j$.
Then, it uses these centers to compute Silhouette values for testing NetFlows by flagging as OOD those having a maximum value below a threshold.
Note that the simplified Silhouette is adopted here to reduce the computational complexity of the standard Silhouette~\cite{silhouette}.
The second detector is based on the K-Nearest Neighbor (\texttt{KNN}) proposal in~\cite{knnOOD}, where a separate KNN model is fitted on the embeddings $e_i$ of training classes and used to measure Euclidean distances at inference time.
This detector works similarly to \texttt{SIM}, but it selects the KNN model to query for measuring the distance from the k$^{th}$ nearest neighbor based on the class predicted by the FNN. If such distance is above a threshold, the NetFlow is OOD.
After preliminary analysis, we set $k = 25$ and $\alpha = 100\%$ in all our experiments. We refer the reader to~\cite{knnOOD} for more detailed explanations.

All these detectors rely on thresholds extracted on ID NetFlows, except ODIN and MD, which also require OOD data.
Details on threshold extraction are provided in Sec.~\ref{ssec:details}.
\section{Experimental Setup}
\label{sec:setup}

\subsection{Datasets and Preprocessing}
\label{ssec:preprocess}

\tit{Datasets.} In our experiments, we train models on benign traffic and specific attacks from one dataset. Then, we evaluate such models on remaining attacks from the same dataset as well as attacks from another one.
Thus, we selected two similar labeled datasets:
\textit{IDS2017}~\cite{CICIDS} comprises synthetic traffic and common attacks like DoS (D) and DDoS (DD), while \textit{IDS2018}~\cite{CICIDS} contains more attack variants and is created in a larger network.
You can refer to Tab.~\ref{tab:ensemble} for the list of their attacks.
The traffic of these datasets is transformed into NetFlows with the CICFlowMeter~\cite{FlowMeter}, where each NetFlow is described by a set of more than 80 features.
We purposely chose these datasets as they contain roughly the same attack families and their traffic comes from consecutive years, hence should not differ much. 
By training on some attacks and testing on all the others from both datasets, we can logically simulate all the cases described in Fig.~\ref{fig:spots}.
With a single dataset, it is hard to cover situations like those described in Case 2 of Fig~\ref{fig:spots}, as inducing shifts in known training attacks requires artificial manual crafting of NetFlows.
Contrary, with a dataset comprising the same attacks, we can try to simulate situations of Case 2 without explicit manual intervention.
As an example, we are going to use the D-hulk traffic of IDS2018 for training and test detectors on the ``shifted'' D-hulk traffic of IDS2017. 
Lastly, note that solely including more diverse datasets does not help in better modeling Case 2.

\begin{table}
\centering
\caption{The common set of 20 features describing a NetFlow.}
\label{tab:features}
\resizebox{\columnwidth}{!}{
\begin{threeparttable}
    \setlength{\tabcolsep}{4pt}
    \begin{tabular}{|c|l|l|}
        \toprule
            \# & Name & Description \\
        \midrule 
            1 & Dst wk          & Whether destination port is well-known [0, 1023]. \\
            2 & Dst reg         & Whether destination port is registered [1024, 49151]. \\
            3 & Num fwd pkts    & Number of packets outgoing the network. \\
            4 & Num bwd pkts    & Number of packets ingoing the network. \\
            5 & Max fwd pkt     & Maximum size of outgoing packets in the NetFlow. \\
            6 & Max bwd pkt     & Maximum size of ingoing packets in the NetFlow. \\
            7 & Ack cnt         & Number of packets with ACK. \\
            8 & Syn cnt         & Number of packets with SYN. \\
            9 & Rst cnt         & Number of packets with RST. \\
            10 & Duration       & NetFlow duration in seconds. \\
            11 & Pkts/s         & Number of exchanged packets per second. \\
            12 & Fwd pkts/s     & Number of packets outgoing the network per second. \\
            13 & Bwd pkts/s     & Number of packets ingoing the network per second. \\
            14 & Avg IAT        & Average Inter Arrival Time between packets. \\
            15 & Std IAT        & Standard deviations of packet Inter Arrival Times. \\
            18 & Sflow fwd byts & Average number of outgoing packet bytes in sub-flows\tnote{a}. \\
            19 & Sflow bwd byts & Average number of ingoing packet bytes in sub-flows\tnote{a}. \\
            16 & Avg idle       & Average idle time (between sub-flows) of the NetFlow. \\
            17 & Avg active     & Average active time (length of sub-flow) of the NetFlow. \\
            20 & Fwd Seg min    & Minimum segment size in outgoing packets. \\
        \bottomrule
    \end{tabular}
    \begin{tablenotes}
        \item[a] A sub-flow is a sequence of packets inside the NetFlow each received within a maximal inter-arrival time.
    \end{tablenotes}
\end{threeparttable}
}
\end{table}
\tit{Preprocessing.} We have established with a simple feature selection procedure a common set of 20 features for both our datasets from the 80+ generated by the CICFlowMeter.
Before applying our procedure, we log-scale continuous features, leave unaltered integer ones, and encode in one-hot port numbers by considering three intervals: well-known, registered, and ephemeral ports.
Then, our feature selection procedure starts by considering each dataset per se and identifies the most important features with a Random Forest analysis~\cite{ImportanceRF}.
On each dataset, we apply the following steps:
\begin{enumerate}
    \item Remove IPs and quasi-constant (variance $<$0.05) features.
    \item Keep an arbitrary feature between ones having a Pearson correlation coefficient higher than 0.8.
    \item Fit a large Random Forest (200 trees with 20 as maximum depth) on the remaining features and evaluate Gini and Permutation importance~\cite{ImportanceRF}.
    \item Rank the features based on the normalized sum of Gini and Permutation importance.
\end{enumerate}
After having the features ranked by their importance, our procedure automatically selects those that are among the 20 most important in both datasets.
To ensure a satisfactory detection performance, we additionally ensure that the top-7 features on each dataset are selected.
The final set of features is reported in Tab.~\ref{tab:features}.
Note that for open-source datasets not having all our features, Zeek with a customized script can be used to generate the required NetFlow features.

\subsection{Model, Training, and Tuning details}
\label{ssec:details}
\tit{Architecture.} In our FNN, we use linear layers of decreasing size in the encoder, the first contains 128 neurons, the second 64, the third 32, and the fourth 2.
All dropout layers switch off neurons with a probability of 0.3.
Whereas the classifier contains as many neurons as classes in the training set. 
Note that we restrict the encoder to produce embeddings in a 2D space to easily plot them. 
We offline verified that this restriction does not limit the model classification performance, as theoretically stated in the universal approximation theorem~\cite{deeplearning}.

\tit{Training.} We train our FNN on scenarios extracted from IDS2018, the larger and more comprehensive dataset, where a scenario comprises all the benign traffic and three attacks (4 classes). Refer to Tab.~\ref{tab:scenarios} for the list of the scenarios and their attacks.
Every training scenario is split 70/30 in a stratified manner. We use 70\% of the traffic for training two separate models -- one with Center-Loss and one with Cross-Entropy. The remaining 30\% is used for validation purposes and detector tuning.
All the models are trained for $25$ epochs with the Adam optimizer~\cite{deeplearning}, batch size of 512, and learning rate at 0.0005. 
The model producing the best F1-score on the validation traffic (always above 99\% in all our scenarios) is saved for testing.
Regarding Center-Loss, we use a separate Adam optimizer, a learning rate of 0.0001, and a weighting factor $\lambda = 1$ (see Sec.~\ref{ssec:contrastive}).
In addition, we use over- and under-sampling to make batches with roughly the same amount of NetFlows for each class.
This is achieved by sampling with repetition a NetFlow with probability inversely proportional to the frequency of its class in the training set.
\begin{table}[t]
    \caption{The training scenarios extracted from IDS2018 and their rationale in our experiments.}
    \label{tab:scenarios}
    \centering
    \resizebox{\columnwidth}{!}{
    \setlength{\tabcolsep}{4pt}
    \begin{tabular}{|c|p{2cm}|p{6cm}|}
        \toprule
            & Training Attacks & Rationale \\
        \midrule
            \multirow{1}{*}{\rotatebox{90}{Scenario 1 \  \ }}
            & 
            \medskip 
            \begin{itemize}
                \item FTP
                \item D-hulk
                \item DD-hoic
            \end{itemize}
            & We select training attacks that generate a high-volume of normal (FTP) and obfuscating traffic. This setup tests detectors in identifying low-volume attacks and variations of training ones, e.g., SSH is a variant of FTP on a distinct protocol and D-hulk from IDS2017 may be a shifted version of 2018 one. \\
        \midrule
            \multirow{1}{*}{\rotatebox{90}{Scenario 2 \;}}
            & \begin{itemize}
                \item SSH
                \item D-hulk
                \item DD-http
            \end{itemize}
            & We choose again training attacks of high-volume which may induce different classification patterns within the FNN compared to Scenario 1. Different patterns can potentially impact the capability of certain detectors to effectively identify unknown attacks. \\
        \midrule
            \multirow{1}{*}{\rotatebox{90}{Scenario 3 \;}}
            & \begin{itemize}
                \item D-eye
                \item DD-udp
                \item Bot
            \end{itemize}
            & We create a diverse set of training attacks, spanning various malicious strategies, mostly relying on the HTTP protocol.
            This allows testing detectors in identifying attacks on different protocols and HTTP attacks with similar or distinct malicious strategies. \\
        \bottomrule
    \end{tabular}}
\end{table}

\tit{Metrics and Detector Tuning.} 
Since our objective is to determine whether unknown malicious traffic can be identified as OOD, we evaluate detectors based on the True Positive Rate (TPR).
In this context, a true positive refers to a NetFlow of an unknown attack labeled as OOD.
We specifically avoid using the F1-score because our detectors are tuned to maintain a low False Positive Rate (FPR) of 5\% on ID traffic. 
However, we do utilize the F1-score when assessing the performance of detector combinations, as the rejected ID traffic may exceed 5\%.
All the detectors selected in Sec.~\ref{ssec:detectors} rely on pre-defined rejection thresholds.
To set such thresholds, we followed a common practice in the literature, ensuring that 95\% of the ID validation traffic (malicious included) is not rejected~\cite{ODIN,GenODIN,unifiedOOD,knnOOD}.
The only exceptions are \texttt{ODIN} and \texttt{MD}, for which we also used OOD attacks to extract the threshold and select $\epsilon \in \{0.0001, 0.001, 0.005, 0.01, 0.05, 0.1, 0.5\}$ with $T=20$.
Specifically, we took advantage of attacks not used in our evaluation, like infiltration and attacks with a few NetFlows, and used them along with validation traffic to tune as in~\cite{ODIN}.
Note that we exclude infiltration as it is not well classified by the FNN nor well detected by OOD techniques with our features, i.e., an example of Case 4 in Fig.~\ref{fig:spots}. 
We also see that using hard-to-discriminate attacks improves the detection capability of \texttt{ODIN} and \texttt{MD}.
For other parameters such as mean and covariance matrix in \texttt{MD}, centers in \texttt{SIM}, and \texttt{KNN} models, we extracted them from the training set.
\section{Results \& Analysis}\label{sec:results}

In this section, we evaluate the chosen detectors in Sec.~\ref{ssec:detectors} to identify previously unknown intrusions as OOD.
Remember that these detectors work beside the ML-based NIDS, i.e., the FNN described in Sec.~\ref{ssec:model}, that is trained to classify attacks of specific scenarios.
These scenarios comprise only a few training attacks and are designed to test detectors under different circumstances, such as those presented in Fig.~\ref{fig:spots}. 
The specific scenarios adopted are outlined in Tab.~\ref{tab:scenarios}. 
Although it is hard to precisely pinpoint which situation of Fig.~\ref{fig:spots} is in each scenario, we tried to design them to logically contain all. 
Every scenario comprises training attacks characterized by distinctive aspects.
Within the pool of unknown testing attacks, encompassing all attacks not encountered during training, there are fairly similar and dissimilar ones. 
These testing attacks should end up in different regions of the FNN's decision space, effectively simulating the situations depicted in Fig.~\ref{fig:spots}.

\subsection{Detecting Unknown Attacks with OOD detectors}
\label{ssec:res}

We begin by assessing detectors and their combination when applied to models trained in a \emph{multi-class} setting both with and without Center-Loss (CL).

\tit{Single Detector Results.} 
We first examine the performance of individual detectors.
Tab.~\ref{tab:ood_res} presents in each horizontal section the TPR of detectors (columns) on a distinct scenario. 
Each cell contains the TPR of a detector on an unknown attack when applied to the FNN trained with and without Center Loss (TPR$_{CL}\,/\,\text{TPR}_{CE}$).
The last row of a section (Total TPR) reports the global TPR, irrespective of the attack types. 
\begin{table}[t]
    \caption{The True Positive Rate Percentage of OOD detectors when applied to models trained with and without Center Loss.}
    \label{tab:ood_res}
    \centering
    \resizebox{\columnwidth}{!}{
    \setlength{\tabcolsep}{3pt}
    \begin{tabular}{|cl|cccccc|}
        \toprule
            \multicolumn{2}{|c|}{IDS2018} & \multicolumn{6}{c|}{Scenario 1: FTP - D-hulk - DD-hoic} \\
            \multicolumn{2}{|c|}{Attacks} & {\footnotesize \texttt{CONF}} & {\footnotesize \texttt{MCD}} & {\footnotesize \texttt{ODIN}} & {\footnotesize \texttt{MD}} & {\footnotesize \texttt{KNN}} & {\footnotesize \texttt{SIM}} \\
        \midrule
            \multirow{8}{*}{\rotatebox{90}{\footnotesize unknown attacks}} 
             & {\footnotesize SSH}      & 50.0\,/\,50.0  & 53.2\,/\,70.9 & 50.0\,/\,48.2  & 50.0\,/\,48.2 & 100\,/\,100   & 50.0\,/\,48.2 \\
             & {\footnotesize D-eye}    & 95.7\,/\,81.5  & 98.7\,/\,98.7 & 17.1\,/\,61.7  & 92.7\,/\,60.2 & 98.6\,/\,85.9 & 75.2\,/\,60.6 \\
             & {\footnotesize D-http}   & 9.4\,/\,9.4    & 8.1\,/\,7.3   & 0.0\,/\,100    & 100\,/\,100   & 1.0\,/\,1.0   & 2.0\,/\,2.0 \\
             & {\footnotesize D-loris}  & 8.3\,/\,0.4    & 52.9\,/\,26.6 & 8.3\,/\,0.0    & 53.4\,/\,0.4\;  & 79.1\,/\,78.3 & 1.3\,/\,0.6 \\
             & {\footnotesize Web}      & 21.2\,/\,33.4  & 35.4\,/\,32.5 & 3.8\,/\,0.0    & 49.6\,/\,2.5\;  & 56.9\,/\,46.8 & 47.3\,/\,2.3 \\
             & {\footnotesize Botnet}      & 0.0\,/\,0.0    & 0.8\,/\,0.1   & 49.9\,/\,0.0   & 0.4\,/\,0.1   & 50.0\,/\,97.0 & 0.0\,/\,0.0 \\
             & {\footnotesize DD-http}  & 89.4\,/\,48.7  & 95.6\,/\,71.3 & 90.6\,/\,0.0   & 56.0\,/\,0.1  & 93.9\,/\,92.3 & 50.5\,/\,0.0 \\
             & {\footnotesize DD-udp}   & 0.0\,/\,0.0    & 98.1\,/\,0.8  & 0.0\,/\,0.0    & 100\,/\,99.0  & 100\,/\,100   & 100\,/\,0.0 \\
        \midrule
            \multicolumn{2}{|c|}{Total TPR} &   \textbf{53.2}\,/\,33.9    &   \textbf{57.3}\,/\,48.1    &   \textbf{61.6}\,/\,20.6    &   \textbf{48.5}\,/\,20.7    &   74.2\,/\,\textbf{83.9}    &   \textbf{33.8}\,/\,9.5    \\
        \midrule \midrule
             \multicolumn{2}{|c|}{IDS2018} & \multicolumn{6}{c|}{Scenario 2: SSH - D-hulk - DD-http} \\
             \multicolumn{2}{|c|}{Attacks} & {\footnotesize \texttt{CONF}} & {\footnotesize \texttt{MCD}} & {\footnotesize \texttt{ODIN}} & {\footnotesize \texttt{MD}} & {\footnotesize \texttt{KNN}} & {\footnotesize \texttt{SIM}} \\
        \midrule
        \multirow{8}{*}{\rotatebox{90}{\footnotesize unknown attacks}}
             & {\footnotesize FTP}      & 100\,/\,100    & 100\,/\,100   & 99.2\,/\,100  & 100\,/\,100   & 100\,/\,100   & 100\,/\,100 \\
             & {\footnotesize D-eye}    & 66.5\,/\,73.2  & 83.9\,/\,100  & 19.5\,/\,52.7 & 95.9\,/\,70.4 & 100\,/\,100   & 76.3\,/\,7.5 \\
             & {\footnotesize D-http}   & 100\,/\,100    & 100\,/\,100   & 98.1\,/\,100  & 100\,/\,100   & 100\,/\,100   & 100\,/\,100 \\
             & {\footnotesize D-loris}  & 30.5\,/\,0.4\;   & 72.0\,/\,11.5 & 6.8\,/\,0.4   & 77.5\,/\,0.4  & 60.8\,/\,80.9 & 54.7\,/\,0.3 \\
             & {\footnotesize Web}      & 9.4\,/\,10.1   & 57.4\,/\,29.5 & 2.8\,/\,0.9   & 53.7\,/\,3.1  & 40.5\,/\,55.3 & 19.2\,/\,0.0 \\
             & {\footnotesize Botnet}      & 0.0\,/\,0.0    & 0.5\,/\,0.0   & 0.0\,/\,0.0   & 0.3\,/\,0.1   & 48.1\,/\,99.4 & 0.0\,/\,0.0 \\
             & {\footnotesize DD-udp}   & 0.0\,/\,0.0    & 74.1\,/\,59.7 & 14.4\,/\,0.1  & 99.5\,/\,100  & 100\,/\,100   & 99.0\,/\,0.6 \\
             & {\footnotesize DD-hoic}  & 76.1\,/\,76.1  & 76.4\,/\,74.1 & 64.2\,/\,50.2 & 73.0\,/\,10.9 & 60.4\,/\,20.2 & 57.9\,/\,1.4 \\
        \midrule
             \multicolumn{2}{|c|}{Total TPR} &   \textbf{65.1}\,/\,\textbf{65.1}    &   \textbf{66.4}\,/\,65.1    &  \textbf{57.2}\,/\,51.4   &     \textbf{65.1}\,/\,32.3     & \textbf{68.8}\,/\,59.5    &    \textbf{56.6}\,/\,25.4          \\
        \midrule \midrule
             \multicolumn{2}{|c|}{IDS2018} & \multicolumn{6}{c|}{Scenario 3: D-eye - Bot - DD-udp} \\
             \multicolumn{2}{|c|}{Attacks} & {\footnotesize \texttt{CONF}} & {\footnotesize \texttt{MCD}} & {\footnotesize \texttt{ODIN}} & {\footnotesize \texttt{MD}} & {\footnotesize \texttt{KNN}} & {\footnotesize \texttt{SIM}} \\
        \midrule
        \multirow{8}{*}{\rotatebox{90}{\footnotesize unknown attacks}}
             & {\footnotesize FTP}      & 0.0\,/\,0.0    & 7.2\,/\,2.8   & 0.0\,/\,0.0    & 90.0\,/\,89.1   & 94.7\,/\,100   & 78.4\,/\,0.0 \\
             & {\footnotesize SSH}      & 0.0\,/\,0.0     & 10.3\,/\,1.5 & 0.0\,/\,0.0    & 0.0\,/\,1.3   & 0.1\,/\,60.0   & 0.0\,/\,0.0 \\
             & {\footnotesize D-hulk}   & 100\,/\,97.5    & 100\,/\,99.6   & 98.4\,/\,97.3  & 98.4\,/\,96.1   & 99.0\,/\,100   & 91.0\,/\,72.6 \\
             & {\footnotesize D-http}   & 0.0\,/\,0.0         & 7.3\,/\,2.4   & 0.0\,/\,0.0   & 89.8\,/\,88.3   & 99.5\,/\,100   & 45.1\,/\,0.0 \\
             & {\footnotesize D-loris}  & 66.5\,/\,45.5   & 67.0\,/\,55.7 & 45.7\,/\,46.1   & 88.7\,/\,49.7  & 65.8\,/\,80.6 & 3.8\,/\,1.9 \\
             & {\footnotesize Web}      & 3.6\,/\,28.6   & 11.4\,/\,23.8 & 1.0\,/\,9.2    & 53.7\,/\,29.1  & 37.5\,/\,60.1 & 29.2\,/\,2.1 \\
             & {\footnotesize DD-http}  & 91.7\,/\,49.2  & 72.7\,/\,65.7 & 46.3\,/\,46.0   & 44.2\,/\,0.1\;  & 80.0\,/\,70.5 & 43.8\,/\,3.5 \\
             & {\footnotesize DD-hoic}  & 73.1\,/\,94.0  & 61.3\,/\,96.2 & 47.4\,/\,87.0  & 0.0\,/\,0.9\;   & 4.8\,/\,20.5 & 0.0\,/\,7.1 \\
        \midrule
             \multicolumn{2}{|c|}{Total TPR} &    \textbf{66.4}\,/\,61.3   &   59.9\,/\,\textbf{67.2}    &    46.6\,/\,\textbf{58.3}   &    \textbf{45.2}\,/\,33.4   &   56.7\,/\,\textbf{64.8}    &    \textbf{39.4}\,/\,17.9   \\
        \bottomrule
    \end{tabular}}
\end{table}

Overall, we observe that all the unknown attacks are detected to some extent in their traffic.
The best OOD detector appears to be \texttt{KNN}, followed by \texttt{MCD} and \texttt{CONF}, while other detectors exhibit lower average performance.
Specifically, we see that \texttt{ODIN} and \texttt{MD} have generally lower TPRs than confidence-based detectors, contrary to what was discovered in computer vision~\cite{ODIN,GenODIN,unifiedOOD}. 
This suggests that controlled perturbations rigidly derived from the gradient do not always benefit detection as expected. We suspect that NetFlow features, which do not have a bounded and discrete domain as pixels, may require more flexible per-feature perturbations that better conform to the domain of features. This might help in pushing ID NetFlows toward their class, better enlarging the gap between ID and OOD scores as in computer vision.

Furthermore, we find that applying CL to multi-class models does not always improve OOD detection.
Although the embeddings produced with CL are in general more discriminative, this benefits detectors such as \texttt{CONF}, \texttt{MD}, and \texttt{SIM}, but not as much \texttt{KNN}.
Our explanation is that a multi-class FNN has already semantically rich embeddings, reducing the effect of CL. In addition, training with CL may sometimes force the model to produce embeddings closer to known classes, which would not be without CL (refer to the right of Fig.~\ref{fig:tech} for a graphical comparison).
This may benefit the assumptions of certain detectors like the representativeness of mean and covariance matrix in \texttt{MD} and the centers of \texttt{SIM}.
However, tighter neighborhoods may negatively impact the performance of \texttt{KNN} in certain situations.
Consequently, we conclude that CL is a simple method to enhance OOD detection in NIDSs~\cite{knnOOD,cade}, although its effectiveness may vary on certain detectors.

\tit{Ensembles Results.} We proceed by aggregating detectors into two ensembles to improve overall performance and evaluate their complementarity.
For this analysis, we use the previously considered scenarios and assess the ensembles' performance on unknown attacks also from the IDS2017 dataset.

To measure the maximum amount of unknown traffic that can be rejected, we use a simple ensemble (ENS$_1$) that flags a NetFlow as OOD if at least one detector predicts it as such.
This ensemble comprises all the detectors applied to the FNN trained with and without CL, resulting in a total of 12 combinations.
The second ensemble (ENS$_2$) consists of three detectors and flags a NetFlow as OOD if at least one predicts OOD.
We use the \texttt{CONF} detector applied to the CL-trained FNN, along with the \texttt{KNN} and \texttt{ODIN} detectors applied to the FNN trained with Cross-Entropy.
The goal of ENS$_2$ is to prove that it contains complementary detectors.
We remark that these ensembles have been specifically designed to increase the detection (TPR) of unknown attacks.

\begin{table}[t]
    \caption{The True Positive Rate percentage of the OOD ensembles. Cells marked with * contain attacks seen in training.}
    \label{tab:ensemble}
    \centering
    \resizebox{.85\columnwidth}{!}{
    \setlength{\tabcolsep}{3pt}
    \begin{tabular}{|cll|cc|cc|cc|}
        \toprule
            & & & \multicolumn{2}{c|}{Scenario 1} & \multicolumn{2}{c|}{Scenario 2} & \multicolumn{2}{c|}{Scenario 3} \\
            & Attacks & Support & {\footnotesize ENS$_1$} & {\footnotesize ENS$_2$} & {\footnotesize ENS$_1$} & {\footnotesize ENS$_2$} & {\footnotesize ENS$_1$} & {\footnotesize ENS$_2$} \\
        \midrule
            \multirow{13}{*}{\rotatebox{90}{IDS2018}} 
            & {\footnotesize FTP} & 193.4k     & *      & *       & 100     & 100   & 100   & 100 \\
            & {\footnotesize SSH} & 187.6k     & 100    & 100     & *       & *     & 62.0  & 59.1 \\
            & {\footnotesize D-eye} & 41.5k    & 100    & 100     & 100     & 100   & *     & * \\
            & {\footnotesize D-hulk} & 461.9k  & *      & *       & *       & *     & 100   & 100 \\
            & {\footnotesize D-http} & 139.9k  & 100    & 100     & 100     & 100   & 100   & 100 \\
            & {\footnotesize D-loris} & 11.0k  & 100    & 79.0    & 100     & 100   & 100   & 80.7 \\
            & {\footnotesize Web} & 833        & 98.1   & 65.5    & 82.8    & 60.0  & 96.4  & 66.0 \\
            & {\footnotesize Botnet} & 286.2k  & 99.7   & 99.5    & 99.3    & 99.2  & *     & * \\
            & {\footnotesize DD-http} & 576.3k & 97.5   & 94.1    & *       &    *  & 99.8  & 99.5 \\
            & {\footnotesize DD-udp} & 1728    & 100    & 100     & 100     & 100   & *     & * \\
            & {\footnotesize DD-hoic} & 686.0k & *      & *       & 82.8    & 81.1  & 97.4  & 96.6 \\
        \midrule
            \multirow{10}{*}{\rotatebox{90}{IDS2017}} 
            & {\footnotesize FTP} & 3967       & 100    & 100  & 100   & 100    & 100    & 100 \\
            & {\footnotesize SSH} & 2976       & 100    & 100  & 100   & 100    & 100    & 100 \\
            & {\footnotesize D-eye} & 7560     & 100    & 95.6 & 100   & 100    & 100    & 100 \\
            & {\footnotesize D-hulk} & 158.3k  & 100    & 89.8 & 100   & 96.7   & 100    & 100 \\
            & {\footnotesize D-http} & 1740    & 100    & 99.3 & 100   & 100    & 100    & 100 \\
            & {\footnotesize D-loris} & 3999   & 100    & 99.8 & 100   & 100    & 100    & 63.1 \\
            & {\footnotesize Botnet} & 736     & 100    & 99.9 & 100   & 100    & 92.2   & 100 \\
            & {\footnotesize PScan} & 159.1k   & 100    & 99.7 & 100   & 100    & 99.8   & 86.9 \\
            & {\footnotesize DD-loit} & 95.1k  & 100    & 99.9 & 100   & 100    & 100    & 100 \\
        \midrule
            \multicolumn{3}{|c|}{Total TPR}   & \textbf{99.1} & 96.7 & \textbf{93.3} & 92.3 & \textbf{96.9} & 95.3 \\
            \multicolumn{3}{|c|}{Total F1}    & 75.9 & \textbf{86.0} & 77.4 & \textbf{82.5} & 73.5 & \textbf{83.5} \\
        \bottomrule
    \end{tabular}}
\end{table}
Tab.~\ref{tab:ensemble} presents the TPR of the ensembles on attacks from the two datasets (horizontal sections).
The last two rows report the total TPR and total F1-Score on all the attacks from both datasets.
Remember that true positives refer to unknown attacks labeled as OOD while false positives correspond to \underline{benign} NetFlows mistakenly marked as OOD.
We exclude training attacks as the FNN detects them correctly.

We first highlight that ENS$_1$ achieves almost perfect TPRs in both datasets, indicating there are complementary detectors in our set.
However, this ensemble strategy also increases the false positives, as remarked by the consistent gap between total TPR and F1-Score in all three scenarios.
In fact, the false positive rate on benign validation traffic goes from 5\% of single detectors (as resulting from the tuning described in Sec.~\ref{ssec:details}) up to 36\% with the ensemble.

On the other hand, ENS$_2$ demonstrates similar total TPRs compared to ENS$_1$ but consistently achieves better F1-Scores.
This improvement is attributed to significantly reduced false positive rates, which are halved compared to those of ENS$_1$. 
We observed that \texttt{CONF} and \texttt{KNN} contribute the most to this ensemble, aligning with the findings in Tab.~\ref{tab:ood_res}, while ODIN gives a smaller nevertheless important contribution.
Overall, ENS$_2$ proves to be a superior ensemble that incorporates complementary detectors.
This highlights the relevance of combining detectors of different natures (e.g., confidence-, distance-, and gradient-based) applied to models trained with different strategies.
By doing this it is possible to fortify defense against the situations described in Fig.~\ref{fig:spots}.

Additionally, we remark that detecting attacks from other datasets appears to be a relatively easier task, indicating experimental bias~\cite{TESSERACT}.
Although we verified the similarity of individual feature distributions between datasets, patterns extracted from IDS2018 differ from those of IDS2017.
This is evident from the almost perfect rejection of IDS2017 attacks, the rejection of attack types included in the training set from the 2018 data (such as FTP and D-hulk in Scenario 1), and the high rejection rates (above 70\%) for IDS2017 benign traffic. 
Note that the benign traffic from IDS2017 is for a different network and is expected that detectors will reject benign traffic incurred on a different network.
In general, there is a need for a methodology that enables better integration of traffic from different networks (datasets) for the purpose of OOD in NIDSs, a topic we will cover in the future.

Finally, we also conducted experiments by training on IDS2017 attacks and testing on IDS2018.
In these regards, we only observed lower detection rates on certain unknown attacks of IDS2017 like Bot in both individual detectors and ensembles.
However, the performance on attacks from IDS2018 (Bot included) was almost perfect.
This discrepancy vouches once again for the necessity of a better integration methodology.
Due to space limitations, we do not report such extensive results.

\subsection{Better Embedding, Better Detection}
\label{ssec:multi-class}

Many supervised datasets in network intrusion detection contain information about the specific type of attack each NetFlow belongs to. 
Typically, this information is ignored as the task is treated as a binary classification one.
However, we do demonstrate herein that leveraging the richer semantics of multi-class models can improve OOD detection and that Contrastive Learning can serve a similar goal.
To this end, we compare the overall \emph{multi-class} results from the previous section with those of the same detectors applied to models trained in a \emph{binary task}, which is obtained by grouping NetFlows of training attacks into a single malicious class.
We retrain a binary FNN with and without CL for each scenario of Tab.~\ref{tab:scenarios}, and evaluate OOD detectors on unknown attacks.

\tit{Single Detector Comparison.}
We first compare the results of individual detectors on unknown attacks of IDS2018.
In Tab.~\ref{tab:ood_bin}, the top section displays the total TPR of detectors applied to binary models, with and without CL (TPR$_{CL}\,/\,\text{TPR}_{CE}$).
Whereas the bottom section presents the reduction in the rejections computed by subtracting the multi-class total TPRs from binary ones, with and without CL. 
\begin{table}[t]
    \caption{The True Positive Rate Percentage of detectors when applied to binary models trained with and without Center-Loss and the difference with multi-class total TPRs of Tab.~\ref{tab:ood_res}.}
    \label{tab:ood_bin}
    \centering
    \resizebox{\columnwidth}{!}{
    \setlength{\tabcolsep}{2pt}
    \begin{tabular}{|c|cccccc|}
        \toprule
            IDS2018 & {\footnotesize \texttt{CONF}} & {\footnotesize \texttt{MCD}} & {\footnotesize \texttt{ODIN}} & {\footnotesize \texttt{MD}} & {\footnotesize \texttt{KNN}} & {\footnotesize \texttt{SIM}} \\
        \midrule
            {\footnotesize Scenario 1} &   \textbf{54.1}\,\,/\,\,8.9    &   \textbf{61.0}\,/\,47.8    &   \textbf{57.0}\,/\,17.0    &   \textbf{52.6}\,/\,15.9    &   \textbf{85.0}\,/\,73.3    &   \textbf{53.2}\,/\,13.1 \\
            {\footnotesize Scenario 2} &   \textbf{65.4}\,/\,64.4    &   \textbf{66.3}\,/\,65.0    &   \textbf{63.9}\,/\,61.0    &   \textbf{55.3}\,/\,30.8    &   \textbf{57.9}\,/\,50.2    &   \textbf{47.4}\,/\,1.5  \\
            {\footnotesize Scenario 3} &   \textbf{51.3}\,/\,36.1    &   \textbf{52.7}\,/\,46.3    &   \textbf{51.3}\,/\,35.6    &   \textbf{48.5}\,/\,34.0    &   \textbf{53.1}\,/\,52.6    &   \textbf{33.4}\,/\,13.9  \\
        \midrule \midrule
            {\footnotesize $\Delta$Scenario 1} & \up0.9\,/\,\low25.0 & \up3.7\,/\,\low0.3 & \low4.6\,/\,\low3.6 & \up4.1\,/\,\low4.8 & \up10.8\,/\,\low10.6 & \up19.4\,/\,\up 3.6 \\
            {\footnotesize $\Delta$Scenario 2} & \up0.3\,/\,\low0.7 & \low0.1\,/\,\low0.1 & \up3.9\,/\,\up9.6 & \low9.8\,/\,\low1.5 & \low10.9\,/\,\low9.3 & \low9.2\,/\,\low23.9 \\
            {\footnotesize $\Delta$Scenario 3} & \low15.1\,/\,\low25.2 & \low7.2\,/\,\low20.9 & \up4.7\,/\,\low22.7 & \up3.3\,/\,\up0.6 & \low3.6\,/\,\low12.2 & \low6.0\,/\,\low4.0 \\
        \bottomrule
    \end{tabular}}
\end{table}

The results in the top section highlight that CL consistently improves all the TPRs in the binary case, resulting more effective than in multi-class settings.
Therefore, the importance of CL and Contrastive Learning is more pronounced for ML-based NIDSs trained on binary tasks, since multi-class trainings already make embeddings more discriminative.

In the bottom section, we generally observe that detectors applied to the binary FNN \underline{without CL} have significantly reduced \low\ TPRs compared to the multi-class case. 
This underlines that semantic information induced by multi-class training enables to make better embedding spaces for detecting unknown attacks.
Whereas detectors applied to the binary FNN \underline{with CL} have either similar detection rates or less pronounced reductions, suggesting that CL roughly gives the same enhancement despite the training regimes of the model.

Therefore, we conclude that better embeddings, such as those obtained from multi-class models or Contrastive Learning methods, enhance OOD detection.

\begin{figure}[t]
    \centering
    \includegraphics[width=.85\columnwidth]{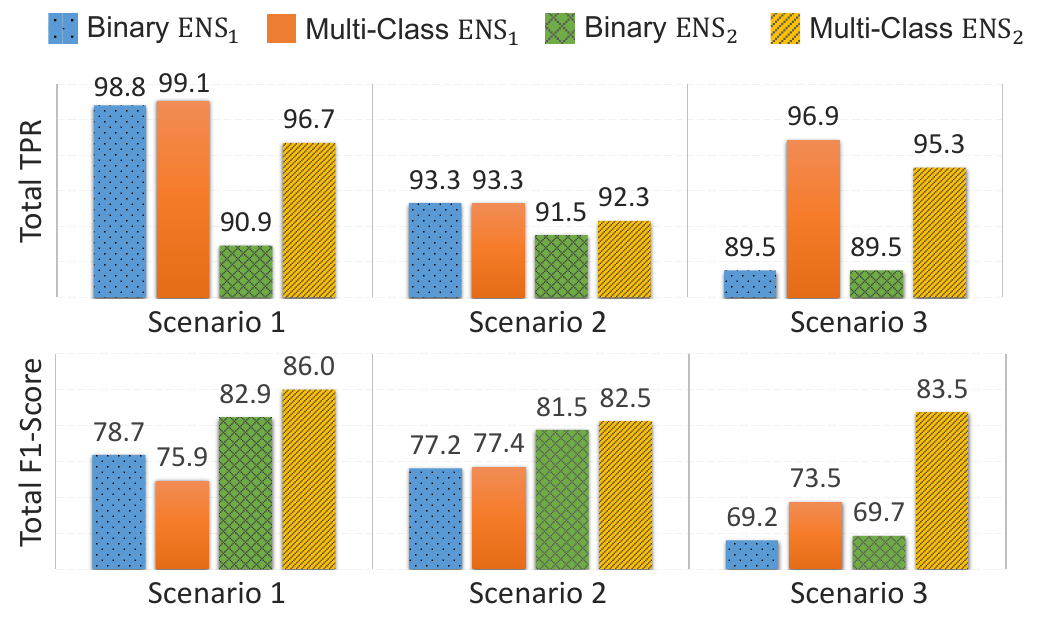}
    \caption{The overall comparison of the four ensembles on unknown attacks from both datasets. Binary ENS$_1$ and ENS$_2$ are those created with detectors applied to the FNN trained in a binary task (benign vs. malicious), while Multi-Class ones are those created with the FNN trained in a multi-class setting (each training attack makes a distinct class). Better viewed in colors.}
    \label{fig:ensembles}
\end{figure}

\tit{Ensemble Comparison.}
Lastly, we present the overall results of the two ensembles described in Sec.~\ref{ssec:res} in the binary case.
Remember that ENS$_1$ comprises all the combinations of binary models and detectors, while ENS$_2$ combines \texttt{CONF} with the CL-trained FNN, as well as \texttt{KNN} and \texttt{ODIN} coupled with the FNN trained without CL.
For this comparison, we consider the two ensembles made with the binary FNNs and also those created with the multi-class FNNs.
Fig.~\ref{fig:ensembles} plots for each scenario of Tab.~\ref{tab:scenarios} the total TPR and F1-Score on unknown attacks from both IDS2018 and IDS2017.

Overall, we observe that ensembles of detectors applied to binary models still yield superior detection, but not as much as in the multi-class case.
This suggests that combining OOD detectors is more effective when applied to models with semantically richer embeddings, such as those produced in multi-class settings.
Furthermore, the better F1-scores of ENS$_2$ with respect to ENS$_1$ in both the binary and multi-class cases indicate that ENS$_2$ detectors positively complement each other.
This demonstrates again the importance of leveraging OOD detectors of different natures, as they enable a broader coverage of unusual and potentially harmful regions of the model's decision space (see Fig.~\ref{fig:spots}).
\section{Conclusion}\label{sec:conclusion}

In this work, we analyzed the ability of existing OOD techniques to detect traffic of unknown intrusions.
We use a standard FeedForward Neural Network as ML-based NIDS and trained it on subsets of attacks in a binary and multi-class setting, by also applying a Contrastive Learning signal.
Then, we use these models along with a set of six OOD techniques relying on different strategies to identify unknown attacks extracted from the same and a separate dataset (network).

Our findings reveal that existing OOD detectors constitute a valid means to identify portions of unknown attacks, although their effectiveness varies compared to other ML fields.
Furthermore, we highlighted that employing training strategies such as multi-class supervision and Contrastive Learning improves the performance of most tested OOD detectors.
Lastly, we demonstrated that combining detectors relying on different strategies leads to superior performance, especially when applied to differently trained models.

While our study has provided some insights into the potential of adopting OOD techniques for network intrusion detection, we acknowledge that there is still much to cover. Notably, one of the limitations of our work is the lack of a methodology that allows a more realistic integration of unknown attacks extracted from diverse datasets (networks). As many datasets offer only limited coverage of cyberattacks, this methodology is of utmost importance to comprehensively assess OOD techniques. 
Additionally, we recognize the prospective value of a visualization tool derived from our plotting strategy used for Fig.~\ref{fig:tech} to inspect models' decision spaces. Such a tool could prove beneficial for network inspection in practical use cases and aid the categorization of attacks in the context of Fig.~\ref{fig:spots}.

Therefore, in future works, we will focus on these points and also explore the influence of different features on the efficacy of OOD detectors.
Furthermore, we intend to improve less effective detectors, like ODIN and MD, and evaluate others.

\bibliographystyle{IEEEtran}
\bibliography{main}

\begin{thebibliography}{10}
\providecommand{\url}[1]{#1}
\csname url@samestyle\endcsname
\providecommand{\newblock}{\relax}
\providecommand{\bibinfo}[2]{#2}
\providecommand{\BIBentrySTDinterwordspacing}{\spaceskip=0pt\relax}
\providecommand{\BIBentryALTinterwordstretchfactor}{4}
\providecommand{\BIBentryALTinterwordspacing}{\spaceskip=\fontdimen2\font plus
\BIBentryALTinterwordstretchfactor\fontdimen3\font minus
  \fontdimen4\font\relax}
\providecommand{\BIBforeignlanguage}[2]{{%
\expandafter\ifx\csname l@#1\endcsname\relax
\typeout{** WARNING: IEEEtran.bst: No hyphenation pattern has been}%
\typeout{** loaded for the language `#1'. Using the pattern for}%
\typeout{** the default language instead.}%
\else
\language=\csname l@#1\endcsname
\fi
#2}}
\providecommand{\BIBdecl}{\relax}
\BIBdecl

\bibitem{DataNIDS}
D.~Chou and M.~Jiang, ``A survey on data-driven network intrusion detection,''
  \emph{Computing Survey}, vol.~54, no.~9, 2021.

\bibitem{survey2019}
A.~Khraisat, I.~Gondal, P.~Vamplew, and J.~Kamruzzaman, ``Survey of intrusion
  detection systems: techniques, datasets and challenges,'' \emph{Springer
  Cybersecurity}, vol.~2, no.~1, 2019.

\bibitem{flowIDS}
A.~Sperotto, G.~Schaffrath, R.~Sadre, C.~Morariu, A.~Pras, and B.~Stiller, ``An
  overview of ip flow-based intrusion detection,'' \emph{IEEE Communications
  Surveys \& Tutorials}, vol.~12, no.~3, 2010.

\bibitem{NIDSurvey}
S.~Gamage and J.~Samarabandu, ``Deep learning methods in network intrusion
  detection: A survey and an objective comparison,'' \emph{Journal of Network
  and Computer Applications}, vol. 169, 2020.

\bibitem{L2Adapt}
A.~Kuppa and N.~Le-Khac, ``Learn to adapt: Robust drift detection in security
  domain,'' \emph{Computers and Electrical Engineering}, vol. 102, 2022.

\bibitem{ConceptDrift}
J.~Lu, A.~Liu, F.~Dong, F.~Gu, J.~Gama, and G.~Zhang, ``Learning under concept
  drift: A review,'' \emph{IEEE Transactions on Knowledge and Data
  Engineering}, vol.~31, no.~12, 2019.

\bibitem{generalizedOOD}
J.~Yang, K.~Zhou, Y.~Li, and Z.~Liu, ``Generalized out-of-distribution
  detection: A survey,'' \emph{ArXiv}, 2021.

\bibitem{CovariateShift}
J.~Quionero-Candela, M.~Sugiyama, A.~Schwaighofer, and N.~Lawrence,
  \emph{Dataset Shift in Machine Learning}.\hskip 1em plus 0.5em minus
  0.4em\relax The MIT Press, 2009.

\bibitem{adwin}
A.~Bifet and R.~Gavalda, ``Learning from time-changing data with adaptive
  windowing,'' \emph{International Conference on Data Mining}, 2007.

\bibitem{centerloss}
Y.~Wen, K.~Zhang, Z.~Li, and Y.~Qiao, ``A discriminative feature learning
  approach for deep face recognition,'' in \emph{Computer Vision--ECCV}.\hskip
  1em plus 0.5em minus 0.4em\relax Springer International Publishing, 2016.

\bibitem{SurveyAnomaly}
G.~Pang, C.~Shen, L.~Cao, and A.~V.~D. Hengel, ``Deep learning for anomaly
  detection: A review,'' \emph{Computing Survey}, vol.~54, 2021.

\bibitem{unifiedOOD}
M.~Salehi, H.~Mirzaei, D.~Hendrycks, Y.~Li, M.~H. Rohban, and M.~Sabokrou, ``A
  unified survey on anomaly, novelty, open-set, and out of-distribution
  detection: Solutions and future challenges,'' \emph{Transactions on Machine
  Learning Research}, 2022.

\bibitem{OSR}
W.~J. Scheirer, A.~de~Rezende~Rocha, A.~Sapkota, and T.~E. Boult, ``Toward open
  set recognition,'' \emph{IEEE Transactions on Pattern Analysis and Machine
  Intelligence}, vol.~35, no.~7, 2013.

\bibitem{confOOD}
D.~Hendrycks and K.~Gimpel, ``A baseline for detecting misclassified and
  out-of-distribution examples in neural networks,'' \emph{Proceedings of
  International Conference on Learning Representations}, 2017.

\bibitem{ODIN}
S.~Liang, Y.~Li, and R.~Srikant, ``Enhancing the reliability of
  out-of-distribution image detection in neural networks,'' \emph{ArXiv}, 2017.

\bibitem{React}
Y.~Sun, C.~Guo, and Y.~Li, ``React: Out-of-distribution detection with
  rectified activations,'' \emph{Advances in Neural Information Processing
  Systems}, vol.~34, 2021.

\bibitem{MCDropout}
Y.~Gal and Z.~Ghahramani, ``Dropout as a bayesian approximation: Representing
  model uncertainty in deep learning,'' in \emph{Proceedings of the 33rd
  International Conference on Machine Learning}, vol.~48, 2016.

\bibitem{LearningConfidence}
T.~DeVries and G.~W. Taylor, ``Learning confidence for out-of-distribution
  detection in neural networks,'' \emph{ArXiv}, 2018.

\bibitem{GenODIN}
Y.-C. Hsu, Y.~Shen, H.~Jin, and Z.~Kira, ``Generalized odin: Detecting
  out-of-distribution image without learning from out-of-distribution data,''
  in \emph{IEEE Conference on Computer Vision and Pattern Recognition}, 2020.

\bibitem{notConf}
A.~Nguyen, J.~Yosinski, and J.~Clune, ``Deep neural networks are easily fooled:
  High confidence predictions for unrecognizable images,'' in \emph{IEEE
  Conference on Computer Vision and Pattern Recognition}, 2015.

\bibitem{EnoughUQ}
A.~Schwaiger, P.~Sinhamahapatra, J.~Gansloser, and K.~Roscher, ``Is uncertainty
  quantification in deep learning sufficient for out-of-distribution
  detection?'' in \emph{AI Safety (IJCAI)}, 2020.

\bibitem{DensityOOD}
K.~Lee, K.~Lee, H.~Lee, and J.~Shin, ``A simple unified framework for detecting
  out-of-distribution samples and adversarial attacks,'' \emph{Advances in
  Neural Information Processing Systems}, vol.~31, 2018.

\bibitem{DensityRegret}
Z.~Xiao, Q.~Yan, and Y.~Amit, ``Likelihood regret: An out-of-distribution
  detection score for variational auto-encoder,'' in \emph{International
  Conference on Neural Information Processing Systems}, 2020.

\bibitem{knnOOD}
Y.~Sun, Y.~Ming, X.~Zhu, and Y.~Li, ``Out-of-distribution detection with deep
  nearest neighbors,'' in \emph{Proceedings of the 39th International
  Conference on Machine Learning Research}, vol. 162, 2022.

\bibitem{SurveyNIDSAnomaly}
D.~Kwon, H.~Kim, J.~Kim, S.~Suh, I.~Kim, and K.~Kim, ``A survey of deep
  learning-based network anomaly detection,'' \emph{Cluster Computing},
  vol.~22, 2019.

\bibitem{bat}
T.~Su, H.~Sun, J.~Zhu, S.~Wang, and Y.~Li, ``Bat: Deep learning methods on
  network intrusion detection using nsl-kdd dataset,'' \emph{IEEE Access},
  vol.~8, 2020.

\bibitem{cade}
L.~Yang, W.~Guo, Q.~Hao, A.~Ciptadi, A.~Ahmadzadeh, X.~Xing, and G.~Wang,
  ``{CADE}: Detecting and explaining concept drift samples for security
  applications,'' in \emph{30th USENIX Security Symposium}, 2021.

\bibitem{ENIDrift}
X.~Wang, ``Enidrift: A fast and adaptive ensemble system for network intrusion
  detection under real-world drift,'' in \emph{Proceedings of the 38th Annual
  Computer Security Applications Conference}.\hskip 1em plus 0.5em minus
  0.4em\relax ACM, 2022.

\bibitem{INSOMNIA}
G.~Andresini, F.~Pendlebury, F.~Pierazzi, C.~Loglisci, A.~Appice, and
  L.~Cavallaro, ``Insomnia: Towards concept-drift robustness in network
  intrusion detection,'' in \emph{Proceedings of the 14th Workshop on
  Artificial Intelligence and Security}.\hskip 1em plus 0.5em minus 0.4em\relax
  New York, USA: ACM, 2021.

\bibitem{deeplearning}
I.~Goodfellow, Y.~Bengio, and A.~Courville, \emph{Deep learning}.\hskip 1em
  plus 0.5em minus 0.4em\relax MIT, 2016.

\bibitem{kitsune}
Y.~Mirsky, T.~Doitshman, Y.~Elovici, and A.~Shabtai, ``Kitsune: An ensemble of
  autoencoders for online network intrusion detection,'' in \emph{Network and
  Distributed System Security Symposium}, 2018.

\bibitem{ConceptEnsemble}
S.~G. Totad, D.~C. Mulimani, and P.~R. Patil, ``Concept drift adaptation in
  intrusion detection systems using ensemble learning,'' \emph{International
  Journal of Natural Computing Research}, vol.~10, no.~4, 2021.

\bibitem{Calibration}
C.~Guo, G.~Pleiss, Y.~Sun, and K.~Q. Weinberger, ``On calibration of modern
  neural networks,'' in \emph{International Conference on Machine
  Learning}.\hskip 1em plus 0.5em minus 0.4em\relax PMLR, 2017.

\bibitem{constrastive}
A.~Jaiswal, A.~Babu, M.~Zadeh, D.~Banerjee, and F.~Makedon, ``A survey on
  contrastive self-supervised learning,'' \emph{Technologies}, vol.~9, 2021.

\bibitem{LearningRejection}
C.~Cortes, G.~DeSalvo, and M.~Mohri, ``Learning with rejection,'' in
  \emph{International Conference on Algorithmic Learning Theory}, 2016.

\bibitem{supcon}
P.~Khosla, P.~Teterwak, C.~Wang, A.~Sarna, Y.~Tian, P.~Isola, A.~Maschinot,
  C.~Liu, and D.~Krishnan, ``Supervised contrastive learning,'' \emph{Advances
  in Neural Information Processing Systems}, vol.~33, 2020.

\bibitem{issues2017}
G.~Engelen, V.~Rimmer, and W.~Joosen, ``Troubleshooting an intrusion detection
  dataset: the cicids2017 case study,'' in \emph{IEEE Security and Privacy
  Workshops}, 2021.

\bibitem{BaselineOOD}
D.~Hendrycks and K.~Gimpel, ``A baseline for detecting misclassified and
  out-of-distribution examples in neural networks,'' \emph{ArXiv}, 2016.

\bibitem{silhouette}
F.~Wang, H.-H. Franco-Penya, J.~D. Kelleher, J.~Pugh, and R.~Ross, ``An
  analysis of the application of simplified silhouette to the evaluation of
  k-means clustering validity,'' in \emph{Machine Learning and Data Mining in
  Pattern Recognition}.\hskip 1em plus 0.5em minus 0.4em\relax Springer
  International Publishing, 2017.

\bibitem{CICIDS}
I.~Sharafaldin, A.~Lashkari, and A.~Ghorbani, ``Toward generating a new
  intrusion detection dataset and intrusion traffic characterization,''
  \emph{Conference on Information Systems Security and Privacy}, vol.~1, 2018.

\bibitem{FlowMeter}
A.~{Habibi Lashkari}., G.~{Draper Gil}., M.~S.~I. Mamun., and A.~A. Ghorbani.,
  ``Characterization of tor traffic using time based features,'' in
  \emph{Proceedings of the 3rd International Conference on Information Systems
  Security and Privacy}, 2017.

\bibitem{ImportanceRF}
R.~Genuer, J.-M. Poggi, and C.~Tuleau-Malot, ``Variable selection using random
  forests,'' \emph{Pattern Recognition Letters}, vol.~31, no.~14, 2010.

\bibitem{TESSERACT}
F.~Pendlebury, F.~Pierazzi, R.~Jordaney, J.~Kinder, and L.~Cavallaro,
  ``Tesseract: Eliminating experimental bias in malware classification across
  space and time,'' in \emph{Proceedings of the 28th USENIX Conference on
  Security Symposium}, 2019.

\end{thebibliography}

\end{document}